# Intelligent Clinical Documentation: Harnessing Generative AI for Patient-Centric Clinical Note Generation

Anjanava Biswas[1]; Wrick Talukdar[2]
AWS AI & ML, IEEE CIS

**Abstract:-** Comprehensive clinical documentation is crucial for effective healthcare delivery, yet it poses a significant burden on healthcare professionals, leading to burnout, increased medical errors, and compromised patient safety. This paper explores the potential of generative AI (Artificial Intelligence) to streamline the clinical documentation process, specifically focusing on generating SOAP (Subjective, Objective, Assessment, Plan) and BIRP (Behavior, Intervention, Response, Plan) notes. We present a case study demonstrating the application of natural language processing (NLP) and automatic speech recognition (ASR) technologies to transcribe patient-clinician interactions, coupled with advanced prompting techniques to generate draft clinical notes using large language models (LLMs). The study highlights the benefits of this approach, including time savings, improved documentation quality, and enhanced patient-centered care. Additionally, we discuss ethical considerations, such as maintaining patient confidentiality and addressing model biases, underscoring the need for responsible deployment of generative AI in healthcare settings. The findings suggest that generative AI has the potential to revolutionize clinical documentation practices, alleviating administrative burdens and enabling healthcare professionals to focus more on direct patient care.

## I. INTRODUCTION

Clinical documentation is a critical component of healthcare delivery, serving as a comprehensive record of patient encounters, diagnoses, treatment plans, and progress. However, the time-consuming nature of documentation has become a significant burden for healthcare professionals, leading to burnout, medical errors, and compromised patient safety [1,2,3]. With multiple researches finding that physicians and clinicians spending an average of two to three hours per day on documentation tasks [4,5,6] (cite source), there is a pressing need for innovative solutions to streamline this process.

Generative AI, a branch of artificial intelligence focused on generating new content based on training data, holds immense potential for transforming clinical documentation practices. By leveraging natural language processing (NLP) and automatic speech recognition (ASR) technologies, generative AI models can transcribe patient-clinician interactions and generate draft clinical notes, capturing the subjective patient information, objective examination findings, assessments, and treatment plans.

This paper presents a case study exploring the application of generative AI for generating SOAP (Subjective, Objective, Assessment, Plan) and BIRP (Behavior, Intervention, Response, Plan) notes, two widely recognized formats for behavioral health related clinical documentation. We demonstrate the use of advanced prompting techniques to guide large language models (LLMs) in generating comprehensive and structured clinical notes based on transcribed patient-clinician interactions.

We also discuss how these formats of clinical notes can be enhanced and improved to reduce accuracy issues, errors, and improve notes quality throughout the patient's treatment journey via augmenting the system with document data, and additional audio/video data.

## II. PREVIOUS WORK

Several studies have investigated approaches to alleviate the documentation burden on healthcare professionals and improve the accuracy and quality of clinical notes. One line of research has focused on leveraging artificial intelligence (AI) models to generate clinical notes automatically.

A study by Kernberg et al. (2023) [7] evaluated the performance of ChatGPT-4, which is a conversational AI interface built by OpenAI based on GPT-3.5 and GPT-4 large language models, in generating SOAP (Subjective, Objective, Assessment, and Plan) notes based on transcripts of simulated patient-provider encounters. The findings revealed significant variations in errors, accuracy, and note quality produced by the AI model. On average, 23.6 errors per clinical case were identified, with omissions being the predominant type of error, accounting for 86% of the errors. Notably, the accuracy of the generated notes exhibited an inverse relationship with the length of the transcripts and the complexity of the data elements, suggesting potential limitations in handling intricate medical cases. The study concluded that the quality and reliability of AI-generated clinical notes did not meet the standards required for clinical use, highlighting the need for further research to address accuracy, variability, and potential error issues.

Another line of research has explored the use of medical scribes, individuals who accompany healthcare providers during patient encounters and document the interactions in real-time. Rule et al. (2022) [8] conducted a retrospective cross-sectional study analyzing over 50,000 outpatient progress notes, some written with scribe assistance and others without. The study revealed that scribed notes were consistently longer than those written without scribe assistance, with much of the additional content originating from note templates. Furthermore, scribed notes were more likely to include certain templated lists, such as the patient's medications or past medical history. However, the study also observed significant variations in how working with scribes affected each provider's documentation workflow, suggesting that providers adapt their note-taking practices to varying degrees when assisted by scribes. The findings indicate that while the use of scribes may contribute to note bloat, individual providers' documentation workflows and note templates play a significant role in shaping the contents of scribed notes.

These studies underscore the ongoing efforts to improve clinical documentation practices and highlight the potential benefits, as well as the limitations, of leveraging AI-based solutions and medical scribes to address this challenge. We extend these existing studies, specifically focusing on the ability of AI models to generate accurate, detailed, yet succinct medical notes using commercially available as well as open-source Large Language Models.

## III. METHODOLOGY

The methodology employed in this case study aimed to simulate a real-world healthcare scenario, leveraging cutting-edge technologies to streamline the clinical documentation process. The study followed a systematic approach, encompassing data collection, transcription, prompt engineering, and model selection and deployment. Ethical considerations, such as maintaining patient confidentiality and adhering to guidelines, were of utmost importance throughout the process. This included using research and educational synthetic data available in the public domain.

### A. Data Collection

To obtain authentic patient-clinician interactions, we collaborated with University of Leeds Researcher and Clinical Psychologist Lecturer Judith Johnson and utilized her experimental therapy sessions available on YouTube [9]. These videos featured unscripted dialogues between Johnson and several subjects and mental/behavioral health patients portraying as clients, providing realistic examples of therapeutic encounters. While the sessions were simulated for educational purposes, they accurately depicted the nuances and dynamics of patient-clinician interactions. Appropriate measures were taken to ensure the protection of privacy and confidentiality, as the video content did not contain any personally identifiable information (PII) or protected health information (PHI).

### B. Transcription

The recorded patient-clinician interactions were then processed using state-of-the-art automatic speech recognition (ASR) models, such as OpenAI Whisper [10]. These models were trained on vast datasets to accurately transcribe the audio or video files into text format, capturing the nuances and intricacies of the conversations. The transcription process was crucial for providing the necessary input for the subsequent steps.

While Whisper excels at transcribing audio accurately, it does not perform speaker diarization out of the box. Speaker diarization, the process of separating speech segments by different speakers, is crucial for understanding the context and flow of conversations, especially in multi-speaker scenarios like patient-clinician interactions.

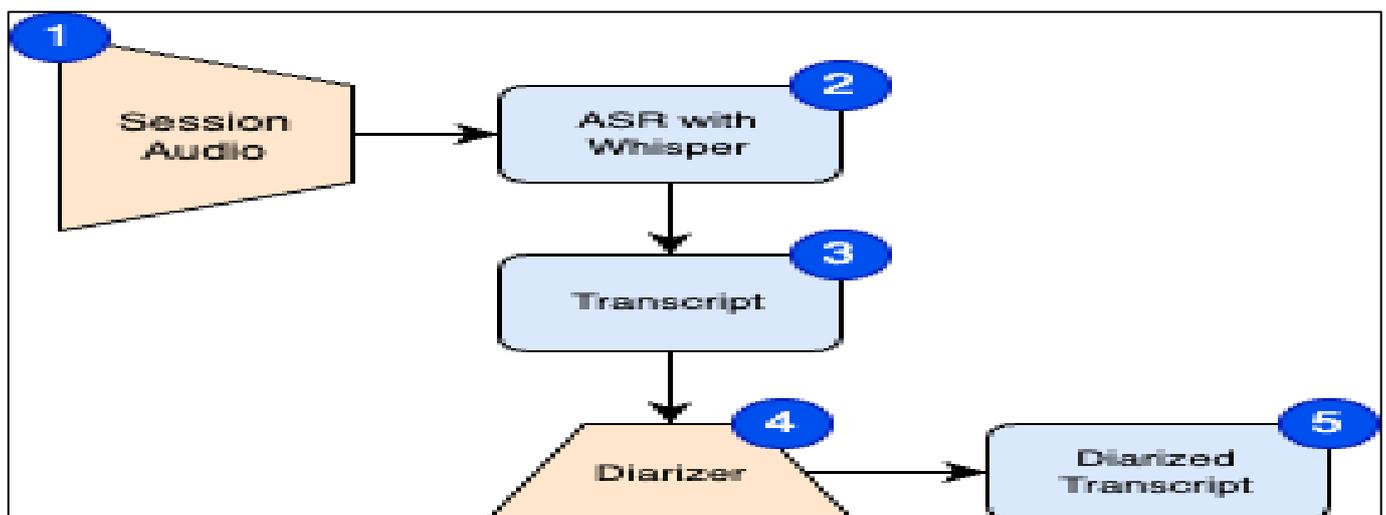

Fig 1: Audio Transcription with ASR and Transcript Diarization

To address this challenge, we analyzed two approaches to diarize the speech with a utterance classification mechanism.

> *Diarization using an Alternate Insanely-Fast-Whisper Model [11]*

This model is an extension of the original Whisper model, specifically designed to perform fast transcription of long audio and comes in with built in support for plugging in diarization model such as pyannotate/speaker-diarization [12]. However, we failed to achieve any significant and successful diarization with the speech audio rendering the result not appropriate for notes generation.

> *Diarization Leveraging GPT-3.5 for Utterance Classification*

In this approach, we first obtained the plain text transcription from Whisper. Then, we utilized GPT-3.5, a powerful language model, to classify each utterance as either spoken by the patient or the clinician. This classification task can be formulated as a binary sequence labeling problem, where each token in the input sequence is assigned a label (0 for clinician, 1 for patient). The probabilities for each class are normalized using the softmax function, which is expressed mathematically as:

$$Softmax(z_i) = \frac{e^{z_i}}{\sum_k e^{z_k}}$$

Here, ( z ) represents the logits or raw outputs from the model's final neural network layer. This normalization ensures that the predicted probabilities are distributed over the two classes, facilitating a clear classification.

To measure the performance of our model, we use the cross-entropy loss function, which quantifies the difference between the predicted probabilities and the actual labels. The loss function is crucial for models dealing with probabilities and is defined as:

$$L = -\sum_{i=1}^{N} (y_i \log(p_i) + (1 - y_i) \log(1 - p_i))$$

Here, ($y_i$) is the true label, and ($p_i$) is the model's predicted probability for each class. The function effectively penalizes the probability divergence from the actual label, driving the model to improve its predictions during training. We also discuss the sequence labeling schema in detail, explaining how the model handles dependencies between labels in a sequence. This aspect is critical since the context within which words or phrases appear can significantly influence their classification.

Table 1: Audio Transcript Generated with Whisper (Non-Diarized) vs. Diarized Audio Transcript with Utterance Classification

| SR Audio Transcript from Whisper Model | |
| --- | --- |
| **Non-Diarized Transcript** | **Diarized Transcript** |
| Hi, Eve. Good to see you again.<br>Hi.<br>Can I take a look at your scores?<br>Sure.<br>So while I'm looking at these, just tell me in your own words how you've been feeling this week.<br>Well, I would say, say I think a little bit better. I don't know if I know exactly why, but I feel a little bit better. Like when I woke up in the morning, I was able to get up more easily. And I think that when I was just like reading the paper, even the sports section, I felt like I was able to concentrate a little better.<br>Oh, that's wonderful. I'm really glad to hear that. And it looks like you're sleeping better too.<br>Well, I think that what I meant by that was mostly that I didn't oversleep.<br>Okay.<br>Because I had been spending a lot of time in bed and I didn't get up. I mean, I would say I got up at seven, but I didn't really get up at seven. | [0] Hi, Eve. Good to see you again.<br>[1] Hi.<br>[0] Can I take a look at your scores?<br>[1] Sure.<br>[0] So while I'm looking at these, just tell me in your own words how you've been feeling this week.<br>[1] Well, I would say, say I think a little bit better. I don't know if I know exactly why, but I feel a little bit better. Like when I woke up in the morning, I was able to get up more easily. And I think that when I was just like reading the paper, even the sports section, I felt like I was able to concentrate a little better.<br>[0] Oh, that's wonderful. I'm really glad to hear that. And it looks like you're sleeping better too.<br>[1] Well, I think that what I meant by that was mostly that I didn't oversleep.<br>[0] Okay.<br>[1] Because I had been spending a lot of time in bed and I didn't get up. I mean, I would say I got up at seven, but I didn't really get up at seven. |

Additionally, we consider various metrics to evaluate the model's performance, including accuracy, precision, recall, and F1-score. Accuracy is calculated by:

$$\text{Accuracy} = \frac{\text{Number of Correct Predictions}}{\text{Total Number of Predictions}}$$

Precision and recall are particularly useful for assessing performance in scenarios where class distribution is imbalanced, which is often the case in conversational datasets where one party may speak more than the other. These metrics, combined with the confusion matrix, provide a comprehensive evaluation of the model.

### C. Prompt Engineering

To generate structural SOAP and BIRP notes, we considered four large language models for evaluation using diarized speech to text data – GPT-3.5 Turbo, GPT-4 Turbo, Claude V3, and Mixtral8x7b Instruct, and Llama-3 70B Instruct. We employed advanced prompting techniques – with zero shot, and one shot – to guide the models in generating structural SOAP and BIRP notes. With GPT-4 Turbo, we were able to leverage its unique *function calling* feature to further generate programmatically structured SOAP/BIRP notes in JSON format which made it easier to process and consume. For this study, we considered a standard SOAP and BIRP note format (see appendix for detailed format). We employed two main approaches: **basic prompting** and **advanced prompting** techniques.

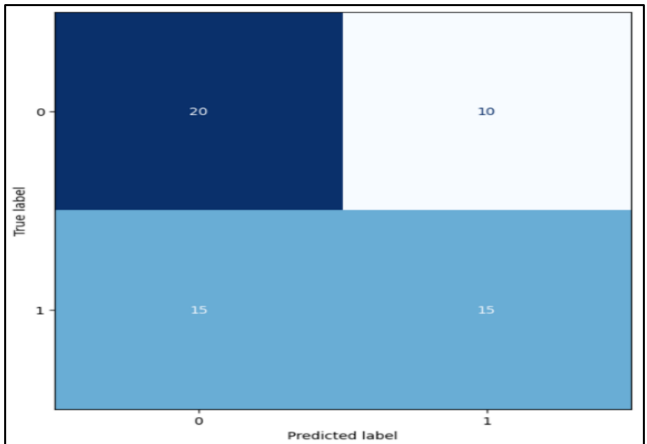

Fig 2: Confusion Matrix of Utterance Classification using Whisper and Pyannotate

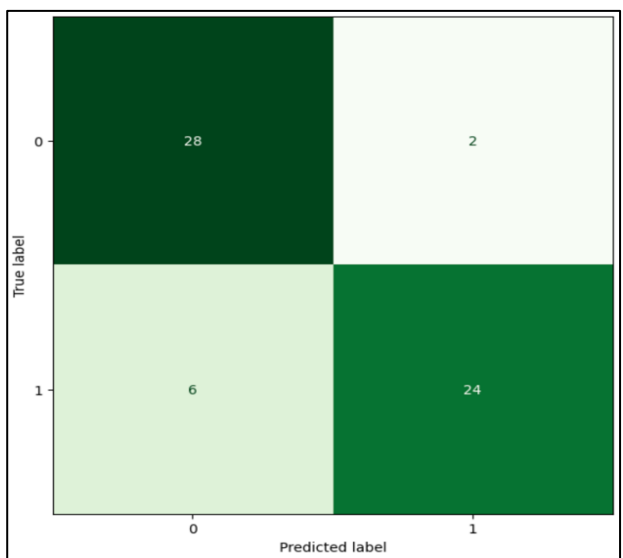

Fig 3: Confusion Matrix of Utterance Classification using Whisper and GPT-3.5

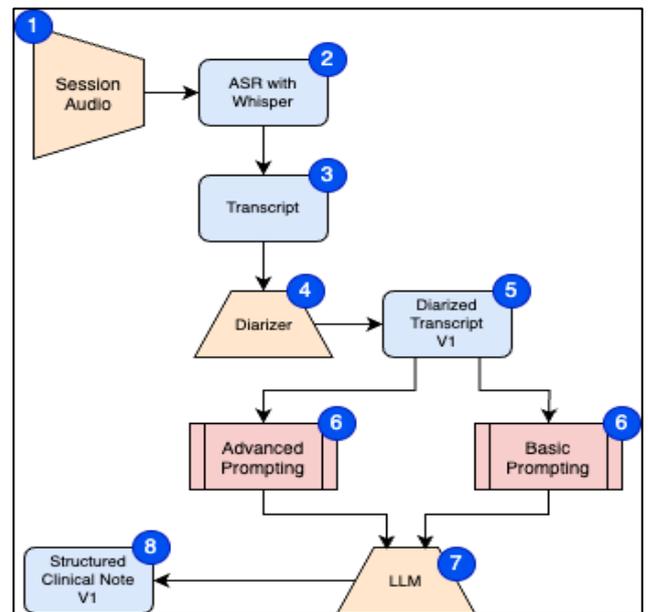

Fig 4: Basic and Advanced Prompting to Generate Structured Clinical Notes

It is worth noting, however, that while this study did not evaluate the effect on the quality of clinical notes due to error rates in classification, such a classification task could potentially be conducted using cheaper and much smaller models such as BERT. These alternatives might offer a more cost-effective solution while still maintaining reasonable accuracy, especially in resource-constrained environments.

➢ *Basic Prompting*

In the basic prompting approach, we provided the diarized transcript as input to the language models, along with instructions to generate SOAP or BIRP notes based on the conversation. This method relied on the model's understanding of the prompt and its ability to extract relevant information from the transcript to construct the clinical note structure. The basic prompting technique served as a baseline to evaluate the models' out-of-the-box performance in generating structured clinical notes.

| Example prompt |
|---|
| • **Transcript**:<br>{{ Diarized transcript of patient-clinician interaction }}<br><br>• **Instructions**: Based on the above transcript, please generate a SOAP note following the Subjective, Objective, Assessment, and Plan format. Include all relevant details from the conversation, and maintain patient confidentiality by avoiding the use of any personally identifiable information. |

The basic prompting technique is analogous to using chat-based interfaces or conversational AI applications like ChatGPT or Claude.ai, where users provide prompts or instructions, and the model generates responses based on its understanding of the input. However, these conversational interfaces often have limitations in terms of the level of control and customization available for the prompts.

While basic prompting allowed the models to generate notes in the desired format, the quality, completeness, and adherence to the specified structure varied significantly across models and transcripts. Some models struggled to capture all the relevant information or organize it correctly within the SOAP or BIRP sections. Additionally, the notes generated through basic prompting often lacked consistency in terms of content organization, level of detail, and overall coherence.

The limitations of basic prompting highlighted the need for more advanced techniques to guide the language models effectively. Factors such as the model's understanding of the prompt, its ability to comprehend the context and nuances of the conversation, and its capacity to structure information coherently played crucial roles in determining the quality of the generated notes.

Despite its shortcomings, the basic prompting approach served as a valuable starting point, providing insights into the models' inherent capabilities and revealing areas for improvement through more sophisticated prompting techniques.

By directly accessing the models and leveraging advanced prompting techniques, researchers and developers can potentially overcome the limitations of chat-based interfaces and gain greater control over the prompting process. Advanced prompting techniques, such as zero-shot and one-shot learning methods, allow for more explicit guidance and customization, enabling the models to generate more accurate, consistent, and structured outputs. Additionally, direct access to the models may unlock advanced features and capabilities that are not available through conversational interfaces, such as programmatic access, fine-tuning, and integration with other systems or applications.

➢ *Advanced Prompting*

To enhance the quality and consistency of the generated notes, we employed advanced prompting techniques, including zero-shot and one-shot learning methods. These techniques aimed to provide more explicit guidance to the models, leveraging their few-shot learning capabilities.

- **Zero-Shot Prompting** In this approach, we provided detailed instructions related to well-structured SOAP or BIRP notes within the prompt itself, the formatting instructions. The models were expected to understand the desired format using the plain language of the detailed formatting instructions and generate notes accordingly, without any prior fine-tuning or training on similar examples.

---

Example prompt

- **Formatting instructions**: {{ Detailed instructions on the SOAP/BIRP note structure and formatting, including the specific sections and the information to be included in each section }}

- **Transcript**:
  {{ Diarized transcript of patient-clinician interaction }}

- **Instructions**: Based on the above transcript and the provided example, please generate a SOAP/BIRP note following the specified structure and format. Ensure that all relevant information from the transcript is captured in the appropriate sections of the note. Maintain patient confidentiality by avoiding the use of any personally identifiable information.

---

The zero-shot prompting approach relied on the model's ability to understand and generalize from the provided instructions. By giving it detailed instructions related to the structure of a SOAP/BIRP note, we aimed to guide the model in generating notes with a similar level of organization and completeness.

- **One-shot Prompting** Building upon the zero-shot approach, we incorporated a few examples of well-structured SOAP or BIRP notes within the prompt, along with the corresponding transcripts. This method aimed to provide the models with a better understanding of the desired output format and the mapping between the transcript and the generated note.

---

Example prompt

- **Formatting instructions:** {{ Detailed instructions on the SOAP/BIRP note structure and formatting }}
- **Example**: {{ Transcript and corresponding well-structured SOAP/BIRP note }}
- **Transcript**:
  {{ Diarized transcript of patient-clinician interaction }}

---

- **Instructions**: Based on the above examples and the provided transcript, please generate a SOAP/BIRP note following the specified structure and format. Ensure that all relevant information from the transcript is captured in the appropriate sections of the note, while maintaining patient confidentiality by avoiding the use of any personally identifiable information.

By providing multiple examples of well-structured notes and their corresponding transcripts, we aimed to enhance the model's understanding of the desired output format and the relationship between the transcript content and the generated note. This approach leveraged the model's few-shot learning capabilities, allowing it to learn from the provided examples and generalize to new transcripts.

- **Structured prompting** [13]: In addition to zero-shot and one-shot prompting, we explored a third prompting technique that leverages the models' ability to understand structured data formats like JSON (JavaScript Object Notation). This approach involved providing the models with detailed instructions and a JSON schema that defined the structure and fields required for the SOAP or BIRP notes.

JSON Schema is a vocabulary that allows for the annotation and validation of JSON documents. It provides a concise description of the structure and data types expected in a JSON document, enabling the validation of data against the defined schema. By utilizing JSON Schema, we could precisely specify the desired structure and fields for the clinical notes, guiding the models to generate well-formatted and structured outputs.

| Example Prompt |
|---|
| - **Formatting instructions**: {{ Detailed instructions on the SOAP/BIRP note structure and formatting, including the specific sections and the information to be included in each section }}<br>[JSON Schema defining the structure and fields for the SOAP/BIRP note]<br>```
{
    "type": "object",
    "properties": {
        "subjective": {
            "type": "object",
            "description": "Patient's subjective complaints, symptoms, and medical history",
            "properties": {
                "chiefComplaint": {
                    "type": "string"
                },
                "symptoms": {
                    "type": "array",
                    "items": {
                        "type": "string"
                    }
                },
                "medicalHistory": {
                    "type": "string"
                }
            }
        },
        "objective": {...},
        "plan": {...},
        "action": {...},
    }
}
```<br>- **Transcript**:<br>{{ Diarized transcript of patient-clinician interaction }}<br><br>- **Instructions**: Based on the above transcript and the provided example, please generate a SOAP/BIRP note following the specified structure and format. Ensure that all relevant information from the transcript is captured in the appropriate sections of the note. Maintain patient confidentiality by avoiding the use of any personally identifiable information. |

In this prompting technique, we provided the models with a JSON schema that defined the structure and fields required for the SOAP or BIRP note. The schema included a description of each section (subjective, objective, assessment, plan) and the specific fields or properties expected within each section.

For example, the "subjective" section included fields like "chiefComplaint," "symptoms," and "medicalHistory," each with a defined data type (string or array of strings). Similarly, the "objective" section included fields for "vitalSigns," "physicalExamFindings," and "labResults," with nested schemas defining the structure of these fields.

By providing the models with this structured schema, we aimed to guide them in generating notes that strictly adhered to the specified format and included all the required fields. The models were expected to understand the JSON schema and generate a well-structured JSON object representing the clinical note, with the relevant information from the transcript populating the appropriate fields.

This prompting technique leveraged the models' ability to understand and generate structured data formats, allowing for a more precise and controlled generation of clinical notes. Additionally, by using JSON Schema, we could easily validate the generated outputs against the defined schema, ensuring that the notes adhered to the expected structure and field requirements.

In addition to the zero-shot and one-shot prompting techniques, we explored various strategies to optimize the prompts further. These strategies included:

- *Iterative Refinement*: We analyzed the initial outputs generated by the models and used the feedback to refine the prompts, improving clarity and specificity.
- *Prompt Chaining*: [14] We experimented with breaking down the note generation task into smaller subtasks and chaining the prompts together, allowing for a more structured and controlled generation process.
- *Prompt Ensembling*: [15] We explored combining outputs from multiple models using different prompting techniques, leveraging the strengths of each model and prompting approach.





*D. Model Selection and Deployment*

The selection and deployment of the LLM were critical steps in the process. We evaluated various models based on their performance, computational resource requirements, and ethical considerations. Factors such as accuracy, efficiency, and adherence to ethical principles were taken into account. Additionally, we assessed the models' ability to maintain patient confidentiality and avoid the inclusion of identifiable information in the generated notes.

While Mixtral8x7b Instruct and Llama-3 70B Instruct models are open-source models that can be deployed on self-provisioned compute environments, Claude V3 and GPT models are proprietary models accessible only via Anthropic and OpenAI platforms respectively (or via partnering cloud provider platforms) as a hosted model service accessible via API calls.

We considered MMLU [16] (Multitask Multilingual), Narrative QA [17], and MedQA [18] (open domain QA from professional medical board exams) benchmarks for comparison of the four models. Per CRFM (Center for Research on Foundation Models) HELM (Holistic Evaluation of Language Models) framework [19].

- **MMLU**: Claude V3 outperforms all the three models closely followed by GPT-4, Mixtral, and Llama respectively.
- **MedQA**: GPT-4 outperforms all the three models closely followed by Llama, Claude V3 and Mixtral respectively.
- **NarrativeQA**: Llama outperformed all the three models followed by Mixtral, GPT-4, and Claude V3 respectively.

We utilized cloud provider platforms to access these models via Python SDK based API calls. Public Cloud providers have fully managed cloud-based services that make all the four models available to use. This reduced the heavy-lifting of securing compute capacity, and deploy the models in a scalable manner. Additionally, models such as Anthropic Claude V3, and OpenAI GPT-4 are only available via SaaS (Software-as-a-Service) platforms since these are proprietary models.

*E. Comparative Analysis of Models for SOAP and BIRP Notes*

We evaluated across 20 SOAP and BIRP notes, each graded for quality by humans and ranging from simple to complex. We investigated the performance of the four models in SOAP and BIRP notes generation tasks using ROUGE-1 [20] F1 scores. The models assessed include GPT-4, Claude, Llama, and Mixtral, with summaries ranging from basic to complex.

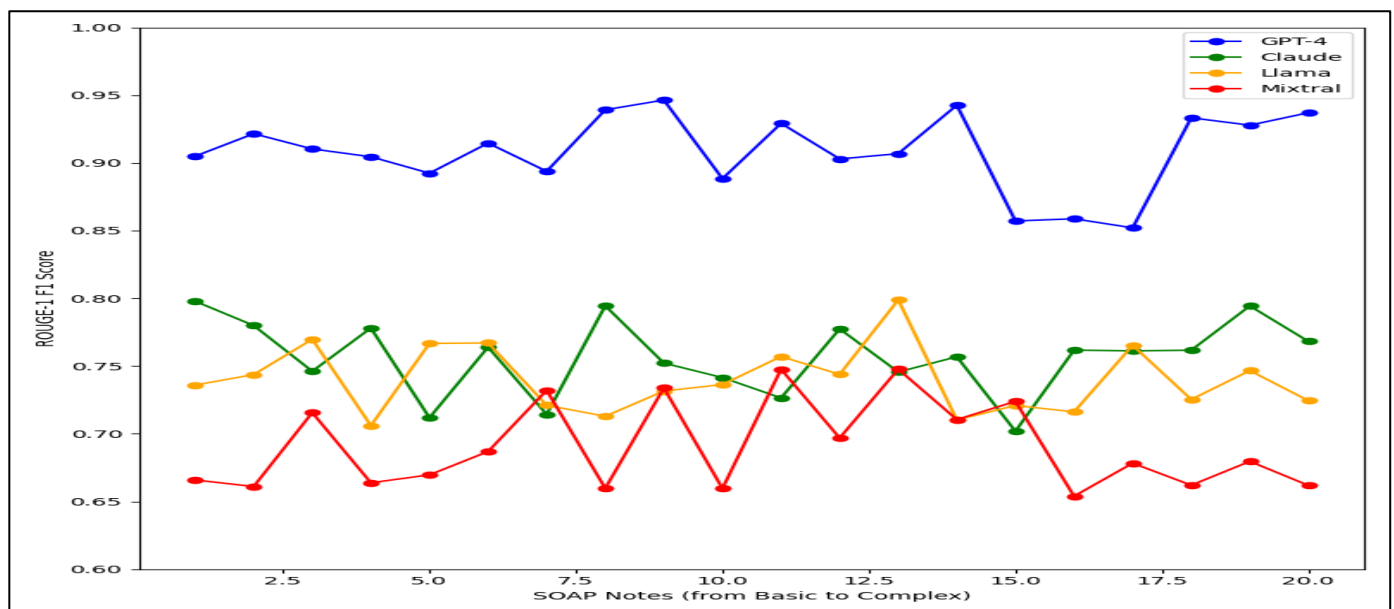

Fig 5: ROUGE-1 F1 Scores for Different Models Across SOAP note Samples

The analysis reveals that GPT-4 consistently achieves superior performance, with ROUGE-1 F1 scores ranging from 0.90 to 0.95. This indicates its robustness and high accuracy across different summary complexities. In contrast, Claude and Llama exhibit similar performance levels, with ROUGE-1 F1 scores fluctuating between 0.70 and 0.80. This suggests moderate stability and adequacy in summarization tasks but highlights a noticeable performance gap compared to GPT-4.





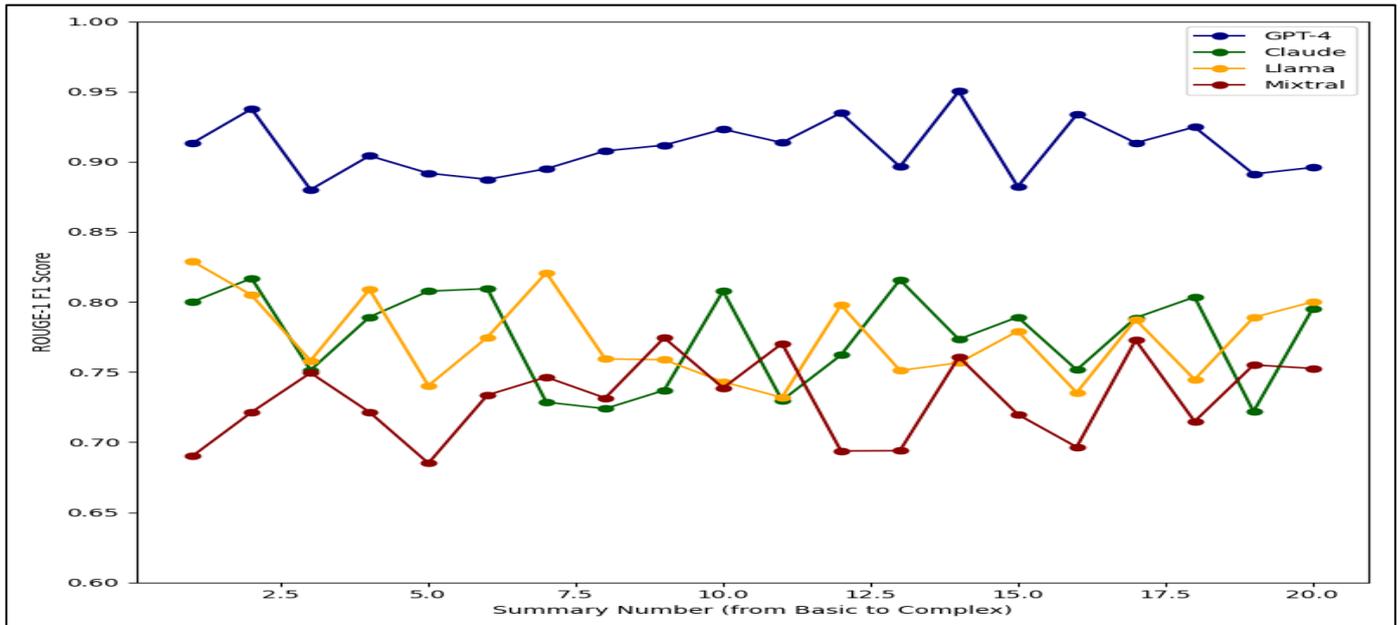

Fig 6: ROUGE-1 F1 Scores for Different Models Across BIRP Note Samples

Mixtral, while trailing behind Claude and Llama, maintains competitive ROUGE-1 F1 scores between 0.65 and 0.75. Despite being the least performant model, Mixtral's scores indicate its ability to generate reasonably accurate summaries, albeit with lower consistency and precision.

## IV. ITERATIVE NOTE IMPROVEMENT FOR EVOLVING PATIENT CARE

Effective healthcare delivery relies on accurate and comprehensive documentation that captures the patient's journey from initial assessment to ongoing treatment and follow-up. However, patient conditions and treatment plans are not static; they evolve over time as new information emerges, and adjustments are made based on the patient's response and progress. This dynamic nature of healthcare necessitates a flexible and adaptive approach to clinical documentation.

In the context of generative AI-powered SOAP and BIRP note generation, the ability to iteratively improve and refine these notes becomes paramount. As patients undergo subsequent clinic visits or encounters, additional data is collected, shedding light on their evolving condition, emerging symptoms, or changes in treatment plans. By harnessing this new information, healthcare providers can ensure that the generated SOAP and BIRP notes remain up-to-date, comprehensive, and reflective of the patient's current state, ultimately enhancing the quality of care and facilitating better clinical decision-making.

*A. Iterative Note Improvement with Subsequent Patient Encounters*

Patient therapy is an ongoing process, and with each visitation or encounter, new information may emerge. This presents an opportunity to refine and improve the generated SOAP and BIRP notes, ensuring that they accurately reflect the patient's evolving condition and treatment plan.

*B. Incorporating Data from Subsequent Encounters*

During each follow-up visit or encounter, additional data can be collected in the form of audio recordings, transcripts, or supplementary documents (e.g., test results, progress reports). This data can be leveraged to enhance the existing SOAP and BIRP notes, making them more comprehensive and up to date.

*C. Incremental Note Generation*

Instead of generating entirely new notes from scratch, the LLMs can be prompted to update and refine the existing notes iteratively, incorporating the new information from subsequent encounters. This approach can be achieved through two main methods:

➢ *Conditional Note Generation*

In this method, the LLM is provided with the existing SOAP or BIRP note, along with the new data (e.g., transcript, audio recording, supplementary documents) from the subsequent encounter. The prompt instructs the model to generate an updated version of the note, considering the previously documented information and incorporating the new relevant details.

| Example prompt |
|---|
| • **Existing SOAP Note**: [Previous SOAP note]<br>• **New Transcript**: [Transcript of the current encounter]<br>• **Instructions**: Based on the existing SOAP note and the new transcript, please generate an updated version of the SOAP note that incorporates relevant information from the current encounter. Maintain the structure and format of the SOAP note, and ensure that all pertinent details from the previous note and the current encounter are accurately reflected. |





➢ *Iterative Note Refinement*

This method involves a multi-step process where the LLM is first prompted to extract the relevant information from the new data (transcript, audio recording, supplementary documents) and then integrate it into the existing SOAP or BIRP note.

- *Step 1*: Extract Relevant Information from the New Data

| Example Prompt |
| --- |
| - **New Transcript:** [Transcript of the current encounter]<br>- **Instructions:** Based on the provided transcript, please extract and summarize the relevant information that should be incorporated into the existing SOAP/BIRP note, such as new symptoms, examination findings, assessments, or treatment plans. |

- *Step 2*: Integrate the Extracted Information into the Existing Note

| Example prompt |
| --- |
| - **Existing SOAP Note**: [Previous SOAP note]<br>- **New Information Summary**: [Summary from Step 1]<br>- **Instructions**: Based on the existing SOAP note and the new information summary, please generate an updated version of the SOAP note that seamlessly incorporates the new relevant details while maintaining the structure and format of the note. |

The overall idea is to iterate on the first version of the clinical note by considering it as the primary context and then augmenting it with further context from subsequent information from documents, or audio.

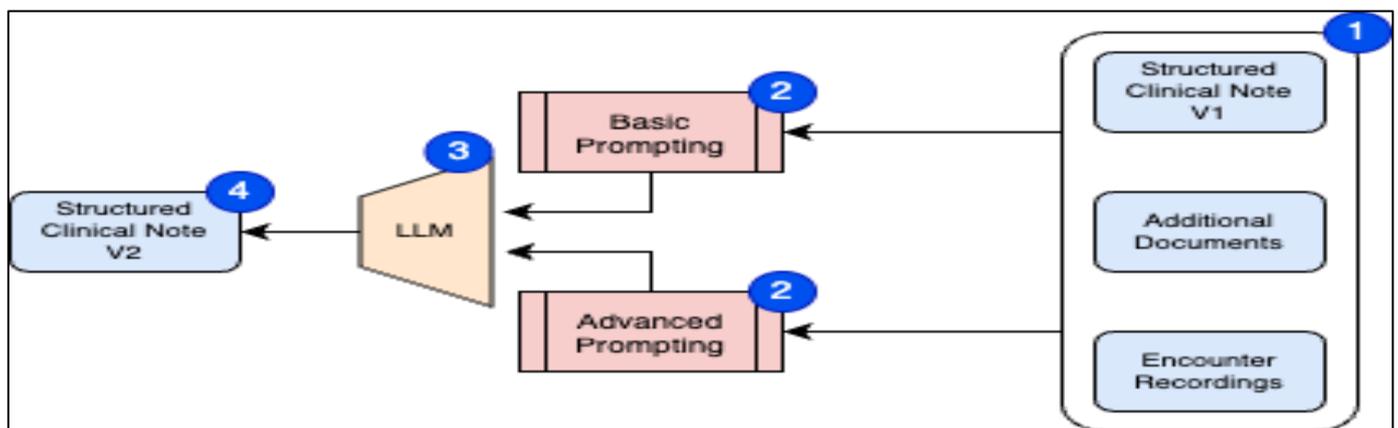

Fig 7: Iterative Clinical Notes Improvement by Augmenting Additional Patient Encounter Specific Data

## D. Continuous Learning and Adaptation

As the iterative note improvement process continues over multiple encounters, the LLMs can continuously learn and adapt to the specific patient's case, capturing the nuances and evolution of their condition and treatment plan. This iterative approach not only enhances the accuracy and completeness of the clinical notes but also facilitates a more personalized and patient-centered approach to care.

## E. Version Control and Auditing

To maintain a comprehensive record of the patient's journey and track the changes made to the SOAP and BIRP notes, it is essential to implement version control and auditing mechanisms. Each iteration of the note can be timestamped and archived, allowing healthcare professionals to review the historical progression of the patient's condition and treatment plan if needed.

By leveraging the iterative note improvement process, healthcare providers can ensure that the SOAP and BIRP notes remain accurate, up-to-date, and reflective of the patient's evolving needs, ultimately enhancing the quality of care and facilitating better clinical decision-making.

## V. CHALLENGES AND FURTHER RESEARCH

While the integration of generative AI in clinical documentation offers numerous benefits, it also presents several challenges that must be addressed to ensure the responsible and effective deployment of this technology in healthcare settings.

## A. Data Quality and Representation

The performance of generative AI models is heavily dependent on the quality and representativeness of the data used for training. In the context of clinical notes generation, the models must be trained on a diverse and comprehensive dataset encompassing a wide range of medical conditions, patient demographics, and clinical scenarios. Failure to do so can lead to biases and inaccuracies in the generated notes, potentially compromising patient care.





Moreover, the healthcare domain is characterized by complex medical terminology, abbreviations, and context-specific language, which can pose challenges for language models to accurately understand and generate relevant content.

*B. Privacy and Security Concerns*

Ensuring the privacy and security of patient data is a critical consideration when leveraging generative AI for clinical documentation. Healthcare organizations must implement robust data protection measures to prevent unauthorized access, data breaches, or unintentional disclosure of sensitive patient information.

Additionally, the generated notes themselves must be carefully scrutinized to ensure that they do not inadvertently include any personally identifiable information (PII) or protected health information (PHI), which could violate privacy regulations and breach patient confidentiality.

*C. Model Interpretability and Transparency*

While generative AI models can produce human-like text, their decision-making processes and reasoning are often opaque, making it challenging to understand and interpret the rationale behind the generated content. In the context of clinical notes, it is crucial for healthcare professionals to understand the basis for the model's assessments, diagnoses, and treatment recommendations to ensure appropriate patient care and mitigate potential errors or biases.

Efforts must be made to enhance model interpretability and transparency, such as through the development of explainable AI (XAI) techniques or the incorporation of domain-specific knowledge and reasoning capabilities into the models.

*D. Model Reliability and Robustness*

The reliability and robustness of generative AI models in generating accurate and consistent clinical notes is a significant challenge. These models may exhibit hallucinations or generate factually incorrect information, which can have severe consequences in healthcare settings.

Rigorous testing and validation processes must be implemented to assess the models' performance across a diverse range of scenarios, including edge cases and rare medical conditions. Additionally, mechanisms for detecting and mitigating potential errors or inconsistencies in the generated notes should be developed to ensure patient safety.

*E. Regulatory Compliance and Liability*

The use of generative AI in clinical documentation must comply with relevant healthcare regulations and guidelines, such as those related to patient privacy, data protection, and medical record-keeping. Failure to adhere to these regulations can result in legal and financial consequences for healthcare organizations.

Furthermore, there are potential liability concerns surrounding the use of AI-generated clinical notes. Determining accountability and responsibility in cases where errors or inaccuracies in the generated notes lead to adverse patient outcomes is a complex legal and ethical issue that requires careful consideration and the development of appropriate risk management strategies.

*F. Human Oversight and Validation*

While generative AI can streamline the clinical documentation process, it is crucial to maintain human oversight and validation. Healthcare professionals must review and verify the accuracy and completeness of the generated notes, ensuring that they align with their clinical judgment and observations.

This human-in-the-loop approach not only enhances patient safety but also facilitates the continuous improvement of the generative AI models through feedback and corrections provided by domain experts.

Addressing these challenges requires a collaborative effort involving technology developers, healthcare professionals, regulatory bodies, and policymakers. By fostering open dialogue, conducting rigorous research, and implementing appropriate safeguards and best practices, the responsible and effective deployment of generative AI in clinical documentation can be achieved, ultimately contributing to improved patient care and healthcare outcomes.

## VI. CONCLUSION

The integration of generative AI in clinical documentation presents a transformative opportunity to streamline the documentation process, alleviate administrative burdens on healthcare professionals, and enhance the overall quality and efficiency of patient care. By leveraging natural language processing, automatic speech recognition, and advanced prompting techniques, generative AI models can transcribe patient-clinician interactions and generate draft clinical notes in structured formats such as SOAP (Subjective, Objective, Assessment, Plan) and BIRP (Behavior, Intervention, Response, Plan).

The case study presented in this paper demonstrates the feasibility and potential benefits of this approach, highlighting the time savings and improved documentation quality achieved through the use of generative AI models. The iterative note improvement process, which incorporates data from subsequent patient encounters, further enhances the accuracy and comprehensiveness of the generated notes, ensuring that they remain up-to-date and reflective of the patient's evolving condition and treatment plan. However, the responsible and effective deployment of generative AI in healthcare settings requires addressing several challenges, including data quality and representation, privacy and security concerns, model interpretability and transparency, model reliability and robustness, regulatory compliance and liability considerations, and the need for human oversight and validation.



Volume 9, Issue 5, May – 2024　　International Journal of Innovative Science and Research Technology
ISSN No:-2456-2165　　https://doi.org/10.38124/ijisrt/IJISRT24MAY1483
Addressing these challenges necessitates a collaborative effort involving technology developers, healthcare professionals, regulatory bodies, and policymakers. By fostering open dialogue, conducting rigorous research, and implementing appropriate safeguards and best practices, the potential benefits of generative AI in clinical documentation can be realized while mitigating potential risks and ensuring patient safety and privacy.

Furthermore, the integration of generative AI in clinical documentation is just the beginning of a broader transformation in healthcare. As this technology continues to evolve and mature, it holds the potential to revolutionize various aspects of healthcare delivery, from personalized treatment planning and decision support to drug discovery and clinical trial design. By embracing the power of generative AI while prioritizing ethical and responsible practices, the healthcare industry can unlock new frontiers in patient care, driving improvements in clinical outcomes, operational efficiency, and overall healthcare quality.

Ultimately, the success of generative AI in healthcare will depend on a delicate balance between technological innovation and an unwavering commitment to patient-centered care, ethical principles, and regulatory compliance. By striking this balance, generative AI can become a catalyst for positive change, transforming the way healthcare is delivered and experienced, and ultimately improving the lives of patients worldwide.

## REFERENCES

[1]. Hall, Louise & Johnson, Judith & Watt, Ian & Tsipa, Anastasia & O'Connor, Daryl. (2016). Healthcare Staff Wellbeing, Burnout, and Patient Safety: A Systematic Review. PloS one. 11. e0159015. 10.1371/journal.pone.0159015.
[2]. Al-Ghunaim T, Johnson J, Biyani CS, Yiasemidou M, O'Connor DB. Burnout and patient safety perceptions among surgeons in the United Kingdom during the early phases of the coronavirus disease 2019 pandemic: A two-wave survey. Scottish Medical Journal. 2023;68(2):41-48.
[3]. Hall LH, Johnson J, Watt I, Tsipa A, O'Connor DB. Healthcare Staff Wellbeing, Burnout, and Patient Safety: A Systematic Review. PLoS One. 2016 Jul 8;11(7):e0159015. doi: 10.1371/journal.pone.0159015. PMID: 27391946; PMCID: PMC4938539.
[4]. Ammenwerth E, Spötl HP. The time needed for clinical documentation versus direct patient care. A work-sampling analysis of physicians' activities. Methods Inf Med. 2009;48(1):84-91. PMID: 19151888.
[5]. [Web][https://blog.hettshow.co.uk/research-reveals-clinicians-spend-a-third-of-working-hours-on-clinical-documentation] Research reveals clinicians spend a third of working hours on clinical documentation.
[6]. [Web][https://buildingbetterhealthcare.com/clinicians-spend-a-third-of-their-time-on-clinical-documentation-204644] Clinicians spend a third of their time on clinical documentation; Jo Makosinski
[7]. Kernberg, Annessa & Gold, Jeffrey & Mohan, Vishnu. (2023). Quality, Accuracy and Reproducibility of Publicly-Available ChatGPT-4-Generated Documentation For Generation of Medical Notes. (Preprint). 10.2196/preprints.54419.
[8]. Rule, Adam & Florig, Sarah & Bedrick, Steven & Mohan, Vishnu & Gold, Jeffrey & Hribar, Michelle. (2022). Comparing Scribed and Non-scribed Outpatient Progress Notes. AMIA ... Annual Symposium proceedings. AMIA Symposium. 2021. 1059-1068.
[9]. [Web][https://www.youtube.com/@JudithJohnsonphd] YouTube Channel; Dr. Judith Johnson
[10]. Radford, Alec & Kim, Jong & Xu, Tao & Brockman, Greg & McLeavey, Christine & Sutskever, Ilya. (2022). Robust Speech Recognition via Large-Scale Weak Supervision.
[11]. [Web] Github [https://github.com/Vaibhavs10/insanely-fast-whisper] Insanely Fast Whisper
[12]. End-to-end speaker segmentation for overlap-aware resegmentation; Bredin et al., 2020;Bredin & Laurent, 2021
[13]. Perot, Vincent, Kai Kang, Florian Luisier, Guolong Su, Xiaoyu Sun, Ramya Sree Boppana, Zilong Wang, Jiaqi Mu, Hao Zhang and Nan Hua. "LMDX: Language Model-based Document Information Extraction and Localization." ArXiv abs/2309.10952 (2023): n. pag.
[14]. Wu, Tongshuang & Jiang, Ellen & Donsbach, Aaron & Gray, Jeff & Molina, Alejandra & Terry, Michael & Cai, Carrie. (2022). PromptChainer: Chaining Large Language Model Prompts through Visual Programming. 1-10. 10.1145/3491101.3519729.
[15]. Pitis, Silviu & Zhang, Michael & Wang, Andrew & Ba, Jimmy. (2023). Boosted Prompt Ensembles for Large Language Models.
[16]. Hendrycks, Dan & Burns, Collin & Basart, Steven & Zou, Andy & Mazeika, Mantas & Song, Dawn & Steinhardt, Jacob. (2020). Measuring Massive Multitask Language Understanding.
[17]. Kočiský, Tomáš & Schwarz, Jonathan & Blunsom, Phil & Dyer, Chris & Hermann, Karl & Melis, Gábor & Grefenstette, Edward. (2017). The NarrativeQA Reading Comprehension Challenge. Transactions of the Association for Computational Linguistics. 6. 10.1162/tacl_a_00023.
[18]. Jin, Di & Pan, Eileen & Oufattole, Nassim & Weng, Wei-Hung & Fang, Hanyi & Szolovits, Peter. (2021). What Disease Does This Patient Have? A Large-Scale Open Domain Question Answering Dataset from Medical Exams. Applied Sciences. 11. 6421. 10.3390/app11146421.
IJISRT24MAY1483　　www.ijisrt.com　　1004

# APPENDIX
# SOAP NOTE FORMAT

A. **SUBJECTIVE**

➢ *Presentation*

- **Chief Complaint**: The client reported experiencing persistent anxiety, difficulty sleeping, and frequent headaches. Quote (Chief Complaint): "I can't seem to relax, and my head always hurts."
- **Impairments and Challenges**: The client described struggles with concentration at work, decreased social interactions, and difficulty managing stress. Their anxiety appeared to contribute to sleep disturbances and chronic tension headaches. Quote (Impairments and Challenges): "I can't focus on anything, and I feel like I'm always on edge."

➢ *Psychological Factors:*

- *Symptom 1:*

✓ Symptom Description: Persistent anxiety and worry throughout the day.
✓ Onset: Gradual, worsened over the past year.
✓ Frequency: Daily.
✓ Ascendance: No improvements reported.
✓ Intensity: Moderate to severe.
✓ Duration: Approximately one year per client report.
✓ Quote (Symptom): "It's like there's a constant weight on my chest."

- *Symptom 2:*

✓ Symptom Description: Difficulty sleeping, including trouble falling and staying asleep.
✓ Onset: Gradual, no specific onset provided.
✓ Frequency: Nightly.
✓ Ascendance: No improvements reported.
✓ Intensity: Moderate.
✓ Duration: Several months per client report.
✓ Quote (Symptom): "I lie awake for hours and then wake up multiple times at night."

B. **OBJECTIVE**

➢ *Clinical Assessment:*

- **Assessment Tool**: Clinical Interview
- **Results**: See above.
- **Status**: Ongoing

➢ *Risk Assessment:*

- **Risks or Safety Concerns**: No immediate risks or safety concerns identified.

➢ *Interventions:*

- **Therapeutic Approach or Modality**: Cognitive-behavioral therapy, mindfulness-based stress reduction Psychological Interventions:

✓ Validated feelings.
✓ Introduced relaxation techniques.
✓ Assigned sleep hygiene practices. Rationale: Reduce anxiety symptoms and improve sleep quality by addressing underlying cognitive and behavioral factors.





C. ASSESSMENT

➤ *Progress and Response:*

- **Response to Treatment**: The client demonstrated some engagement but expressed skepticism about the effectiveness of therapy. Specific Examples or Instances: The client occasionally dismisses the usefulness of relaxation exercises. Quote (Progress): "I don't think these breathing exercises are helping."
- **Challenges to Progress**: Persistent skepticism and resistance to adopting new techniques may hinder progress. Chronic stress and deeply ingrained worry patterns present ongoing challenges. Therapist Observations and Reflections: The client exhibits high levels of stress and may benefit from a more structured approach to learning relaxation techniques. Therapeutic Alliance: The client expressed some reservations but remains open to continued sessions. The therapist acknowledges the client's concerns and emphasizes the importance of consistency.

D. PLAN

➤ *Follow-Up Actions and Plans:*

- **Homework**: Practice assigned relaxation techniques daily and maintain a sleep diary.
- **Plan for Future Session**: Review the sleep diary and relaxation practice log, discuss any obstacles, and introduce progressive muscle relaxation.
- **Plans for Continued Treatmen**t: Continue weekly therapy sessions, consider referral to a sleep specialist if sleep issues persist. **Coordination of Care**: No immediate coordination of care indicated at this time.

**BIRP NOTE FORMAT**

A. BEHAVIOR

➤ *Psychological Factors:*

- *Symptom 1:*

✓ Symptom Description: Depressed mood most of the day, nearly every day.
✓ Onset: Ongoing, no specific onset provided.
✓ Frequency: Daily.
✓ Ascendance: No improvements reported.
✓ Intensity: Moderate to severe.
✓ Duration: Several years per client report.
✓ Quote (Symptom): "I'm always seeing things I know I can't do. It feels heavy."

- *Symptom 2:*

✓ Symptom Description: Low motivation and withdrawal from previously enjoyable activities.
✓ Onset: Gradual, no specific onset provided.
✓ Frequency: Daily.
✓ Ascendance: No improvements reported.
✓ Intensity: Moderate to severe lack of motivation.
✓ Duration: Several years per client report.
✓ Quote (Symptom): "I just go places where there's no one and sit there alone."

➤ *Therapist Observations and Reflections:*
    The client displays cognitive distortions like mental filtering that focus on negative aspects of situations. Increased awareness of these patterns through thought tracking will be beneficial. The client requires support and encouragement to challenge avoidance behaviors.





- *Interventions*
- ✓ Therapeutic Approach Or Modality: Cognitive-behavioral therapy, interpersonal therapy
- ✓ Psychological Interventions:
- o Validated feelings.
- o Encouraged challenge of automatic thoughts.
- o Assigned thought tracking.
- Rationale: Increase awareness of cognitive distortions fueling depression. Begin the process of identifying and challenging automatic negative thoughts.

B. RESPONSE

- *Progress and Response:*

- Response To Treatment: The client displayed limited engagement and motivation for change.
- Specific Examples Or Instances: Client keeps deflecting from talking about certain issues.

- *Quote (Progress): "I don't know, I don't think I can."*

- Challenges To Progress: Lack of motivation and avoidance of social connections will likely impede progress. Negative automatic thoughts and cognitive distortions will also pose a challenge.
- Therapist Observations And Reflections: Client is fused with their negative thoughts, might need to introduce defusion techniques.
- Therapeutic Alliance: The client showed some resistance. She was hesitant to talk about certain things related to her anxiety. The therapist processed that with her.

C. PLAN

- *Follow-Up Actions and Plans:*

- Homework: Complete thought records identifying automatic negative thoughts and labeling cognitive distortions. Engage in one social activity.
- Plan For Future Session: Review thought records, continue cultivating motivation and self-efficacy, begin discussing behavioral activation steps.
- Plans For Continued Treatment: Continue weekly therapy, consider psychiatric referral if lack of progress.
- Coordination Of Care: No coordination of care indicated at this time.